## SOFTWARE METAPAPER

# Picasso: A Modular Framework for Visualizing the Learning Process of Neural Network Image Classifiers

Ryan Henderson and Rasmus Rothe
Merantix GmbH, Leuschnerdamm 31, 10999, Berlin, DE
Corresponding author: Ryan Henderson (ryan@merantix.com)

Picasso is a free open-source (Eclipse Public License) web application written in Python for rendering standard visualizations useful for analyzing convolutional neural networks. Picasso ships with occlusion maps and saliency maps, two visualizations which help reveal issues that evaluation metrics like loss and accuracy might hide: for example, learning a proxy classification task. Picasso works with the Tensorflow deep learning framework, and Keras (when the model can be loaded into the Tensorflow backend). Picasso can be used with minimal configuration by deep learning researchers and engineers alike across various neural network architectures. Adding new visualizations is simple: the user can specify their visualization code and HTML template separately from the application code.

## (1) Overview

### Introduction

Neural networks (NNs) [1] and convolutional neural networks (CNNs) [2, 3, 4] are subject to unique training pitfalls [5, 6]. Consider this motivating example [7]:

> Once upon a time, the US Army wanted to use neural networks to automatically detect camouflaged enemy tanks. The researchers trained a neural net on 50 photos of camouflaged tanks in trees, and 50 photos of trees without tanks[…].
>
> Wisely, the researchers had originally taken 200 photos, 100 photos of tanks and 100 photos of trees. They had used only 50 of each for the training set. The researchers ran the neural network on the remaining 100 photos, and without further training the neural network classified all remaining photos correctly. Success confirmed! The researchers handed the finished work to the Pentagon, which soon handed it back, complaining that in their own tests the neural network did no better than chance at discriminating photos.
>
> It turned out that in the researchers' dataset, photos of camouflaged tanks had been taken on cloudy days, while photos of plain forest had been taken on sunny days. *The neural network had learned to distinguish cloudy days from sunny days, instead of distinguishing camouflaged tanks from empty forest.* [emphasis added]

While this story may be apocryphal, it nonetheless illustrates a common pitfall in machine learning: training on a proxy feature instead of the intended feature. In this case, cloudy vs. sunny instead of tank vs. no tank. As CNNs are increasingly used in critical applications, sound training can literally be a matter of life and death [8].

We developed Picasso to help protect against situations where evaluation metrics like loss and accuracy may not tell the whole story in training neural networks on image classification tasks. Picasso includes two visualizations so far: partial occlusion [10] and saliency mapping [11]. The user may upload new input images and select from among installed visualizations and their attendant settings. Picasso was designed with ease of adding new visualizations in mind, detailed in the Implementation and architecture and Reuse Potential sections. At the time of this writing, Picasso has support for neural networks trained in Keras [12] or Tensorflow [13].

At Merantix, we work with a variety of neural network architectures. Picasso makes it easy to see standard visualizations across our models in various fields: including applications in automotive, such as understanding when road segmentation or object detection fail; advertisement, such as understanding why certain creatives receive higher click-through rates; and medical imaging, such as analyzing what regions in a CT or X-ray image contain irregularities. See **Figure 1** for a screenshot of the Picasso application after computing partial occlusion maps for various images. The user has chosen to use the VGG16 [9] model for image classification. This example is included with Picasso, along with a trained MNIST [14] model in both Keras and Tensorflow.

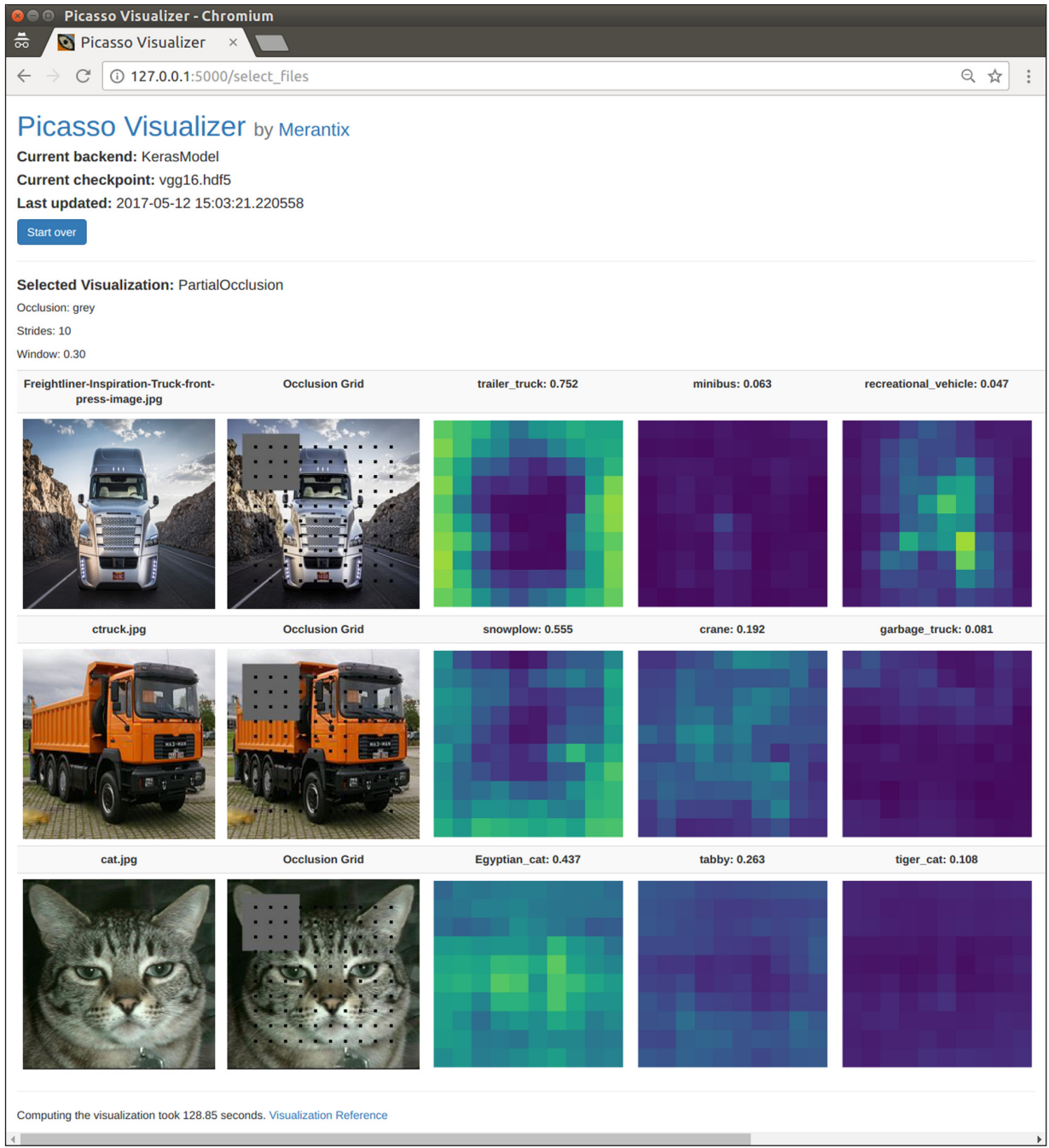

**Figure 1:** A screen capture of the Picasso web application after computing partial occlusion figures for various input images. The classifier is a trained VGG16 [9] network.

Other visualization packages exist to help bring transparency to the learning process, most notably the Deep Visualization Toolbox [15] and `keras-vis` [16], which can also generate saliency maps. There are also various applications for visualizing the computational graph itself and monitoring the evaluation metrics, like Tensorboard. Not all of these tools provide a web application out-of-the-box, however. We furthermore required an application that would easily allow us to add new visualizations, which may in the future include visualizations such as class activation mapping [17, 18] and image segmentation [19, 20].

Let us return to the tank example. Could the visualizations provided with Picasso have helped the Army researchers? We would like to be able to see that our model (VGG16) is classifying based on the "tank-like" features of the image, and not some proxy feature like the weather. See **Figure 2** for the partial occlusion maps generated by Picasso. We see that when we occlude portions of the sky, the model still classifies the image as a tank. Conversely, when we occlude parts of the tank treads, the model is far less certain that the image is a tank.

That the model is classifying on the correct features is further supported by the saliency maps. Saliency maps compute the derivative of the classification for a given class with respect to the input image. Thus regions with high gradient–bright regions–are important to the given classification because changing them would change the classification more relative to other pixels. **Figure 3** shows the saliency map for the tank image. Notice that with a few exceptions the non-tank areas are largely dark, which means changing these pixels should not make the image more or less "tanky."

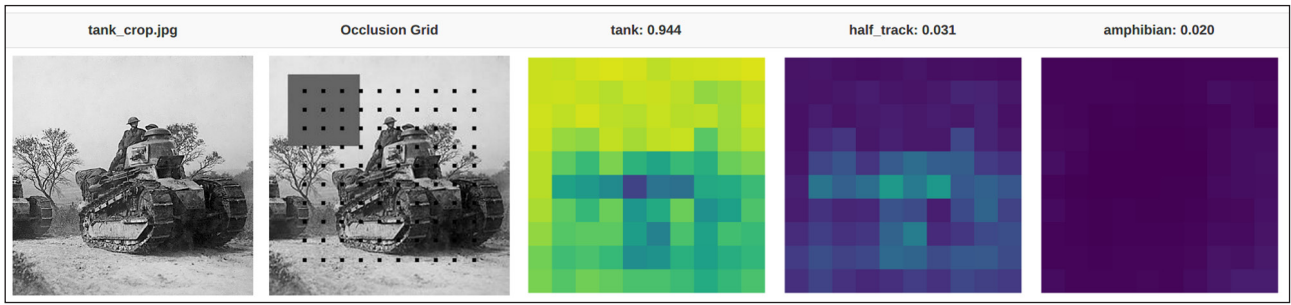

**Figure 2:** The partial occlusion map sequentially blocks out parts of the image to determine which regions are important to classification. The numbers in the header are the overall class probabilities. Brighter regions correspond to areas where the probability of the given class is high–i.e. blocking out this part of the image does not change the classification much. The tank image is in the public domain [21].

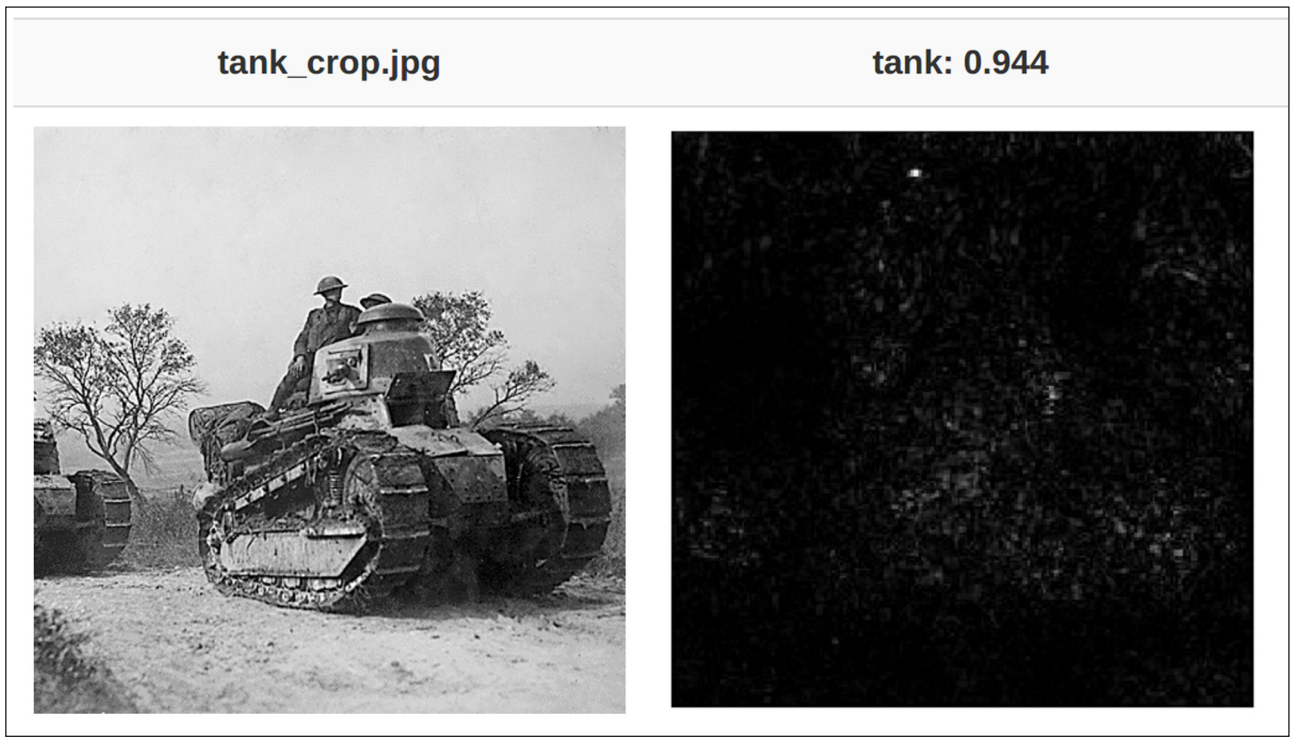

**Figure 3:** Saliency map for the tank. Brighter pixels indicate higher values for the derivative of "tank" with respect to that input pixel for this image. The brightest pixels appear to be in the tank region of the image, which is a good indication the model is classifying on the tank-like features.

**Implementation and architecture**
Picasso was written in Python 3.5 using the Flask web application framework. Visualization classes and HTML templates must be defined separately by the user, but do not require modifying any other source files to use. Picasso handles the uploading of user-supplied images and generates temporary folders containing input and output images. If the visualization class has a `settings` attribute, Picasso automatically renders the settings selection as a separate page.

Application-level settings are handled via a configuration file, where the user may specify the deep learning framework (Keras or Tensorflow) as well as the location of the checkpoint files for their chosen model. The user must also supply a function to preprocess the image (reshape the image into appropriate input dimensions) and decode the output of the model (provide class labels).

**Quality control**
Picasso has unit tests written in the Pytest framework covering the web application functionality, and automatically tests that new visualizations render without errors. The GitHub repository performs continuous integration via Travis-CI. Test coverage is monitored with Codecov. A user can verify the software is working by starting the web application and pointing a web browser to `127.0.0.1:5000`. In addition to docstrings and inline comments, extensive documentation is available on Read the Docs.

## (2) Availability

**Operating system**
Any operating system capable of running Python 3.5 or higher.

**Programming language**
Python >= 3.5.

**Additional system requirements**
None.

**Dependencies**
These Python packages will be installed as part of the normal installation process: click >= 6.7, cycler >= 0.10.0, Flask >= 0.12, h5py >= 2.6.0, itsdangerous >= 0.24, Jinja2 >= 2.9.5, Keras >= 1.2.2, MarkupSafe >= 0.23, matplotlib >= 2.0.0, numpy >= 1.12.0, olefile >= 0.44, packaging >= 16.8, Pillow >= 4.0.0, protobuf >= 3.2.0, pyparsing >= 2.1.10, python-dateutil >= 2.6.0, pytz >= 2016.10, PyYAML >= 3.12, requests >= 2.13.0, scipy >= 0.18.1, six >= 1.10.0, tensorflow >= 1.0.0, Werkzeug >= 0.11.15.

**List of contributors**
- Bunk, Stefan stefan@merantix.com – code review
- Chen, Josh josh@merantix.com – code review
- Henderson, Ryan ryan@merantix.com – code review
- McSpedon, John john@merantix.com – code review
- Rothe, Rasmus rasmus@merantix.com – code review
- Scopel, Filippo filippo@merantix.com – development
- Sprengel, Elias elias@shirp.ch – development

**Software location**
*Archive*
  **Name:** Picasso
  **Persistent identifier:** https://github.com/merantix/picasso/tree/v0.1.1
  **Licence:** EPL
  **Publisher:** Merantix
  **Version published:** v0.1.1
  **Date published:** 12/05/17

*Code repository*
  **Name:** GitHub
  **Persistent identifier:** https://github.com/merantix/picasso
  **Licence:** EPL
  **Date published:** 12/05/17

*Emulation environment*
  **Name:** N/A
  **Persistent identifier:** N/A
  **Licence:** N/A
  **Date published:** N/A

**Language**
English.

## (3) Reuse potential

Any researcher or engineer working with a model in Tensorflow or Keras which takes images as inputs and gives classification probabilities as output can use Picasso with very little effort. Picasso does make some assumptions about the topology of the neural network, but developers can further adapt the Picasso framework to more specialized computational graphs with minimal changes to the code.

Picasso is specifically designed to make implementing new visualizations as painless as possible (see the visualization documentation). New visualization code can be added without modifying any other source code. We hope to add more visualizations as we continue to develop this tool internally, and especially hope for new community-developed visualizations.

**Acknowledgements**
The authors would like to thank the Merantix team for support during development and documentation. Also, thanks to David Dohan and Nader Al-Naji for helpful discussions in preparing this manuscript.

**Competing Interests**
The authors have no competing interests to declare.